\documentclass[conference, a4paper]{IEEEtran}
\IEEEoverridecommandlockouts

	\usepackage[english]{babel}
	\usepackage[utf8]{inputenc} 
	\usepackage[T1]{fontenc}

\usepackage[cmex10]{amsmath}

\usepackage{cite}

\ifCLASSINFOpdf
  \usepackage[pdftex]{graphicx}
\else
\fi
\usepackage[cmex10]{amsmath}

\usepackage{amsmath}
\usepackage{algorithm}
\usepackage[noend]{algpseudocode}

\usepackage{multirow}
\usepackage{array}
\usepackage[lofdepth,lotdepth]{subfig}

\AtBeginDocument{%
  \renewcommand\tablename{TABLO}
}

\AtBeginDocument{%
  \renewcommand\abstractname{Abstract}
}

\begin{document}


%
\title{A Case Study: Using Genetic Algorithm for Job Scheduling Problem }

	\author{\IEEEauthorblockN{Burak Tağtekin, Mahiye Uluyağmur Öztürk and Mert Kutay Sezer}
	\IEEEauthorblockA{Huawei Turkey R\&D Center, Istanbul, Turkey}}



%

\maketitle


\begin{abstract}
Nowadays, DevOps pipelines of huge projects are getting more and more complex.  Each job in the pipeline might need different requirements including specific hardware specifications and dependencies. To achieve minimal makespan, developers always apply as much machines as possible. Consequently, others may be stalled for waiting resource released. Minimizing the makespan of each job using a few resource is a challenging problem.  In this study, it is aimed to 1) automatically determine the priority of jobs to reduce the waiting time in the line, 2) automatically allocate the machine resource to each job. In this work, the problem is formulated as a multi-objective optimization problem. We use GA algorithm to automatically determine job priorities and resource demand for minimizing individual makespan and resource usage. Finally, the experimental results
show that our proposed priority list generation algorithm is more effective than current priority list producing method in the aspects of makespan and allocated machine count.

\end{abstract}



%
\IEEEpeerreviewmaketitle

\IEEEpubidadjcol

\section{Introduction}

Defining a priority list for jobs is a critical issue in DevOps pipelines. Some constraints and requirements should be provided while the priority lists are producing. One of the most important constraint is minimizing the makespan. Makespan is the total time required for jobs to run and finish \cite{Reza}. Another requirement is to minimize the number of machines needed. In the scope of the Job-shop Scheduling Problem (JSP), each jobs include many tasks, these tasks should be run on defined machines and tasks should not be interrupted \cite{paper1}.


JSP requires \textit{n} jobs and \textit{m} machines \cite{paper1}. Each job has varying running time and may need a specific machine to process on it. Also, some jobs may have higher priority than others and need to be run in the first order. JSP aims to create a priority list which promises to minimize the makespan. In addition to minimizing makespan, it is expected that the number of machines allocated will be minimum.


As a solution to JSP we employ a heuristic approach, Genetic Algorithm (GA). The concept of GA is inspired by nature. Weak species are faced with extinction by natural selection \cite{paper2} and strong species are more successful in transferring their genes to subsequent generations via reproduction. In optimization problems, GAs are used to find the strong solution for some scheduling problems and maximum utilization problems \cite{Cormen}. In this work we will employ GA in order to solve a multi-objective problem \cite{Konak}. Multi-objective optimization concerns solving more than one problems simultaneously and in an optimum way. Generally, GA is used for single objective problems. However, by implementing some regulations to fitness function, they can be utilized for multi-objective optimization problems \cite{Yun}.

 In this work, our contribution is instead of manually producing the priority list, we introduce two priority list generating algorithms by using genetic algorithm to automate the process. When producing a priority list, we take into account the dependencies of jobs to each other and the number of machines that jobs needed. The presented model is used to solve a real job scheduling problem in our system. It increased the efficiency by 20\%. We present a unique chromosome representation technique which includes priority list and machine information such as machine count and machine type. Machine type depends on the user input and it is defined before a build runs. On the contrary of \cite{Wu} study, in our work, the machine information part of the chromosome can take unlimited machine type and in the crossover phase each machine type crossed with the same species.

\subsection{Related works} 

Genetic algorithm is used very commonly to solve job or task scheduling problems \cite{Omara}, \cite{Wu}, \cite{Kwok}. There are two type of scheduling methods: static and dynamic. In \cite{Kwok} they solve a static scheduling problem, where job dependencies and machine count are known before execution, similar to us. On the other hand, an adaptive problem representation and a dynamically incremental fitness function are worked on in \cite{Wu}. However, our problem environment is stationary unlike \cite{Wu}.

Main aim of JSP is to minimize the makespan \cite{Xu}. In multi-objective optimization, two or more conflicting objectives are optimized with a given set of constraints simultaneously. Although, in real world problems when an objective optimized this might lead degradation for another objective \cite{Punia}. The first solution to this problem is the vector evaluated GA (VEGA) \cite{Vega} proposed by Schaffer. In literature there are many algorithms which are developed for this purpose. Some of those are; Multi-objective Genetic Algorithm(MOGA) \cite{MOGA}, Niched Pareto Genetic Algorithm(NPGA) \cite{NPGA}, Weight-based Genetic Algorithm (WBGA) \cite{WBGA}.

\section{Genetic Algorithm Details}

\subsection{Fitness function}\label{Fitness Function}
In this work, we need to implement a system infrastructure to check the builds and record the run time of them.

\begin{figure}[h]
	\centering
	\includegraphics[scale=0.7]{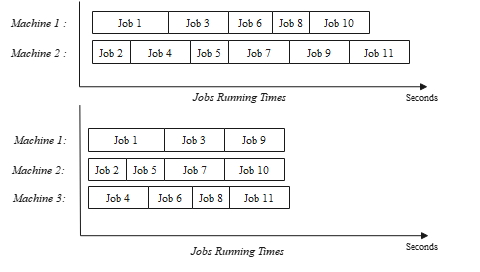}
	\caption{Machine count has effects on build run time.}
	\label{Machine Count Effect}
\end{figure}

As shown in Figure \ref{Machine Count Effect}, increasing the machine count decreases total run time of a build. If a user allocates a lot of machines to finish a build rapidly, this causes long waiting time for other users and starvation \cite{Shousha} in the system. On the other hand, if a user employs less machine for a build, his waiting time will be longer. This is a multi-objective problem.
We used formula of the fitness function as it is seen in Equation 1:

\begin{equation}
	\label{fitness formula}
	\begin{gathered}
 \forall P_i \in {P} \\
{\alpha_P}_i = w_{RT} * P_{MRT} * {{P}_{i}}_{RT} \\
{\beta_P}_i = w_{MC} * P_{MMC} * {{P}_{i}}_{MC} \\
\omega F_{pi} = {\alpha_P}_i + {\beta_P}_i \\
	\end{gathered}
\end{equation}

$P$ stands for population, $P_i$ is individual $i$ of a population, $P_{MRT}$ is the maximum run time of a population's individuals, ${P_i}_{RT}$ is the run time of a individual $i$ in population. $w_{RT}$ is the weight for run time. ${\alpha_P}_i$ shows contribution of run time to the fitness function. $P_{MMC}$ is the maximum machine count of a population's individuals, ${P_i}_{MC}$ is machine count of a individual $i$ in a population. $w_{MC}$ is the weight for machine count. ${\beta_P}_i$ shows contribution of machine count to the fitness function. Summation of ${\alpha_P}_i$ and ${\beta_P}_i$ indicates the fitness value. 

\begin{figure}[h]
	\centering
	
	\includegraphics[scale=0.4]{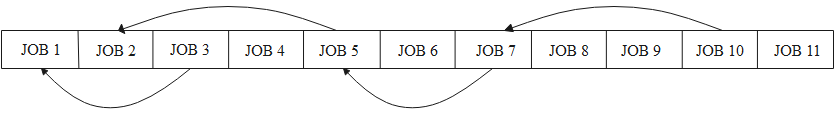}
	
	\caption{An example job queue}
	\label{Job Running Queue}
\end{figure}

Some of the constraints are as follows: 1) every job might run on different operating systems 2) there are some dependencies between the jobs. For example, a job might have to wait for another job to run, in Figure \ref{Job Running Queue} JOB 10 is waiting for JOB 7, JOB 7 is waiting for JOB 5 and so on. We added dependency constraints to the fitness function as can be observed in Algorithm 1. 

\makeatletter
\def\BState{\State\hskip-\ALG@thistlm}
\makeatother

\begin{algorithm}
	\caption{Pseudo code of the fitness function.}\label{Algorithm Fitness}
	\begin{algorithmic}[1]
		\Procedure {Fitness Function}{}
		\State total run time = 0
		\While {priority list not empty}
			\While {machines not full}
				\For{job in priority list}
					\If {job is not waiting dependencies}
						\State {assign the job to a machine}
						\State {remove from priority list}
					\Else 
						\State {continue}
					\EndIf
				\EndFor
			\EndWhile
		
			\State find shortest run time on machines.
			\State decrease all run time with these shortest run time
			\State total run time += shortest run time
		\EndWhile
		\EndProcedure
	\end{algorithmic}
\end{algorithm}

In Algorithm \ref{Algorithm Fitness}, in order to clarify the fitness function to simulate continuous integration (CI) system, we follow these steps: Firstly, the algorithm takes a priority list as an input. From beginning to end, traverse the list and assign the jobs to the available machines. These machines have operating systems such as Windows, Suse, Linux etc. When machines are fully allocated, the algorithm tries to find the job which has the shortest run time. Then shift the time as this shortest run time. Thus, decrease this time value from run time of all jobs like a round robin logic \cite{Rasmussen}. As a result, the machine is ready to run another job. Fill the machine and repeat this algorithm until there is no job on the priority list.

A deadlock situation occurs if a build starts with depended jobs. We propose a deadlock prevention method (see Algorithm \ref{Algorithm One} and Algorithm \ref{Algorithm Two}). Also, since each build has different requirements, they need to run on different operating systems. There are limited types of operating system, but our implementation is generic, it covers all of them. Other important point is that job count in a build might not same for all builds. It depends on development team or project purposes.

\subsection{Chromosome representation}

Build size and machine count of each build might be different. The chromosome should be able to represent all these variations. Therefore, we divide a chromosome into two parts:
\subsubsection{Job queue}
The first part of a chromosome represents a job queue in a build which is called \textit{priority list}. CI system runs the builds according to the priority list.

\subsubsection{Machine count}
The second part of a chromosome includes the machine count. Genetic algorithm finds the most efficient count, but the limit and types of machines depend on the developer.

\begin{figure}[h]
	\centering

	\includegraphics[scale=0.4]{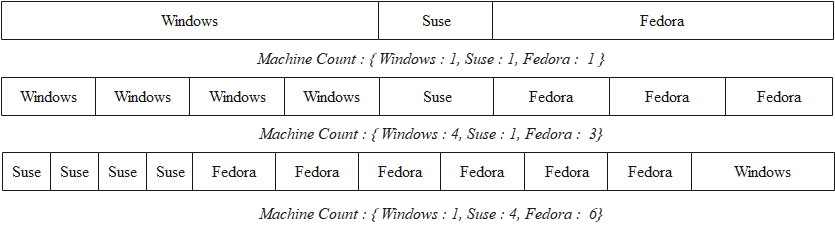}
	
	\caption{Machine types and counts might be different for each build.}
	\label{Types and Counts}
\end{figure}

On the other hand, there are four types of data for each job:
 
\begin{enumerate}
	\item Job Name
	\begin{itemize}
		\item Each job has historical run time values on different builds.
	\end{itemize}
	\item Dependencies
	\begin{itemize}
		\item Dependency array contains name of jobs. It is necessary to prevent the deadlock situation. If the algorithm does not consider the dependency array, it cannot find deadlock on fitness function.
		
	\end{itemize}
	\item Run Time
	\begin{itemize}
		\item Initially zero.
	\end{itemize}
	\item Machine Type
	\begin{itemize}
		\item Windows, Suse, Linux.
	\end{itemize}
\end{enumerate}

On a chromosome, the machine count is defined as binary representation. In Figure \ref{Types and Counts} there are three types of operating systems and also, further operating systems could be added. 

When a new kind of job arrives to the CI system, its run time is not known. To identify total run time of a build, a random run time is assigned to this job. On the other side, if a job has a historical record, our algorithm use it. A job runs in many builds and it might have different run time values because of different environment settings. We chose the quartile value as a run time for this job.

\subsection{Population creation}
In this problem, a population represents priority lists. Initially, it is designed as the number of population is equal to the number of jobs in a build. However, according to early tests, this is not sufficient to reach the best solution. Therefore, we increase the number of population as twice as the number of jobs in a build.

As mentioned in Section \ref{Fitness Function}, the job order is important, because of the deadlock risk. Each candidate priority list must be ready for a smooth transaction. For this reason, the initial population starts with creating suitable priority lists. In this case, if the number of jobs in a build is extremely high, creating the initial population takes too much time.

\makeatletter
\def\BState{\State\hskip-\ALG@thistlm}
\makeatother

\begin{algorithm}
	\caption{First priority list validation algorithm.}\label{Algorithm One}
	\begin{algorithmic}[1]
		\Procedure{Priority list creation}{}
		\While {True}
		\State {shuffle priority list}
		\State {run priority list with fitness function}
		\If {deadlock == true}
		\State \Return true 
		\Else 
		\State \Return priority list 
		\EndIf
		\EndWhile
		\EndProcedure
	\end{algorithmic}
\end{algorithm}

Algorithm \ref{Algorithm One} is effective for small priority lists. If the number of job in a priority list is higher than 200, the running time of the algorithm will takes hours to create a valid population. Hence, we present another algorithm to reduce the running time of the population creation method.

We ensure that there will be no deadlock situation in the second priority list validation algorithm as it is described in Algorithm \ref{Algorithm Two}. Comparing with the Algorithm \ref{Algorithm One}, it is approximately 30 times faster for large builds. This improvement provides us to be able to increase the size and variability of a population.

%

\makeatletter
\def\BState{\State\hskip-\ALG@thistlm}
\makeatother

\begin{algorithm}
	\caption{Second priority list validation algorithm.}\label{Algorithm Two}
	\begin{algorithmic}[1]
		\Procedure{Priority list creation}{}
		\While {True}
		\For {Job in build}
		\State {find indexes of job's each dependencies in priority list}
		\If {job index < biggest dependencies index}
		\State {add job to biggest dependencies index + 1}
		\Else 
		\State \Return continue 
		\EndIf
		\EndFor
		\EndWhile
		\EndProcedure
	\end{algorithmic}
\end{algorithm}

\subsection{Crossover method}

Crossover is the most time consuming part of the genetic algorithm \cite{paper6}. A priority list which has 400 jobs, performs the crossover operation nearly 20.000 times. Every job is unique and the crossover method can change only order of the job list. The size of a priority list does not increase or decrease as a result of crossover. Also, a priority list has all jobs of a build. Therefore, as a crossover method, we need order based crossover algorithms \cite{paper4}. 

Partially mapped crossover (PMX) method, which is an order based crossover technique, is the first alternative \cite{paper5}. It chooses randomly two pivot indexes from chromosomes and changes the parts between two pivot points. A mapping is created for the jobs excluded by pivot points. Then the jobs are swapped regarding to the mapping. PMX creates two individuals instead of one individual as common. This is an important improvement. However, if the size of a priority list is large, the mapping function of PMX consumes too much time. As an alternative method to PMX is Ordered crossover \cite{paper7} that addresses wasting time problem. It is quite simple and fast when comparing with PMX. This algorithm starts same as PMX, but it chooses genes between pivot points from one parent and add them into a child directly on the same position. Ordered crossover fills the empty genes with other parent's genes while preventing repetition. 

\begin{figure}[h]
	\centering
	
	\includegraphics[scale=0.6]{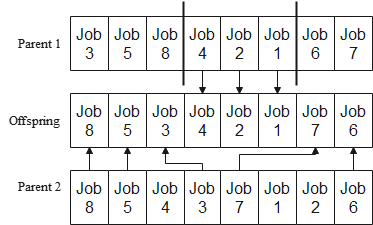}
	
	\caption{Ordered crossover example}
	\label{OX}
\end{figure}

\section{Results}

We present three different test cases to validate the consistency of our algorithm.

\subsubsection{Test Case One}
\begin{figure}
	\centering
	
	\includegraphics[scale=0.3]{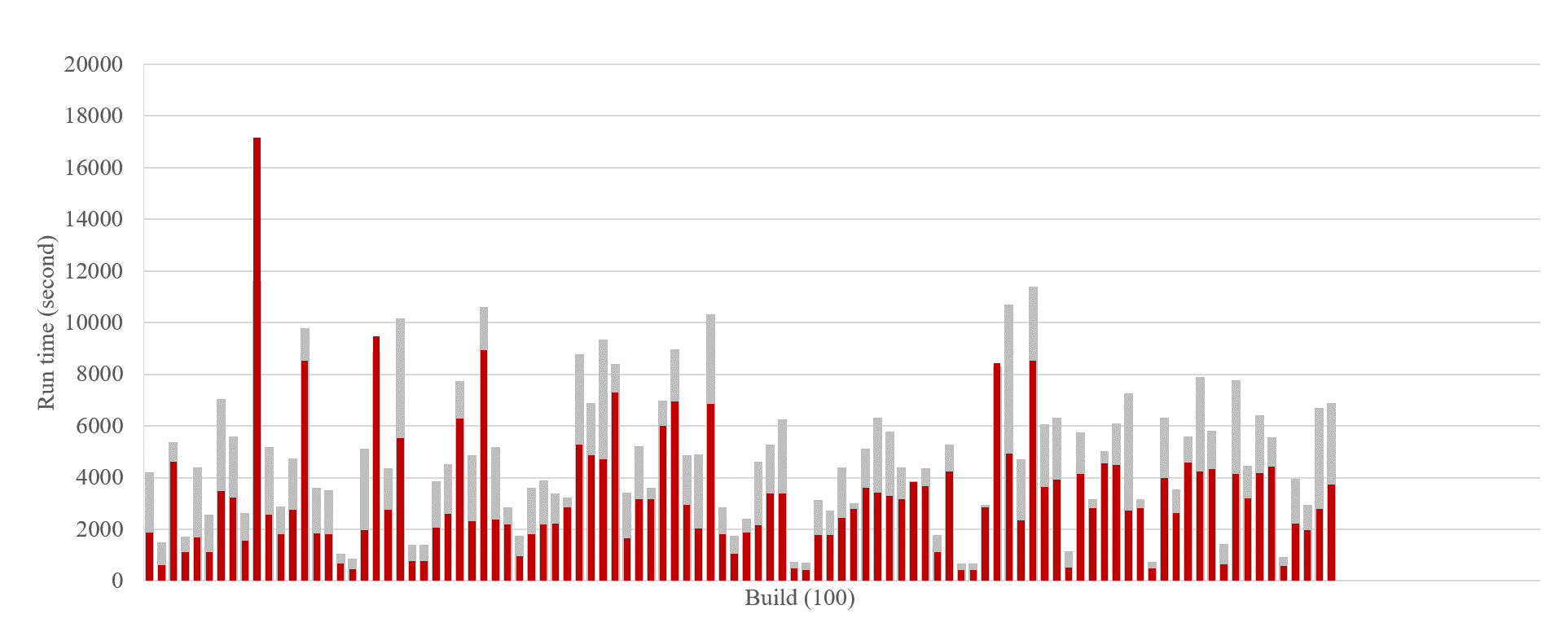}
	\caption{Test case one. Each bar represents a different build. The gray bars show the original run time of the builds and red ones show run time value of the builds with our algorithm. }
	\label{Test Case One}
\end{figure}

For this scenario, we test our algorithm on 100 different builds. Then we calculate the improvement on these builds. As shown in Figure \ref{Test Case One} nearly all red bars are smaller then gray ones. It means our algorithm finds mostly better run time for these builds thanks to produced priority list. 

\subsubsection{Test Case Two}
\begin{figure}
	\centering
	
	\includegraphics[scale=0.25]{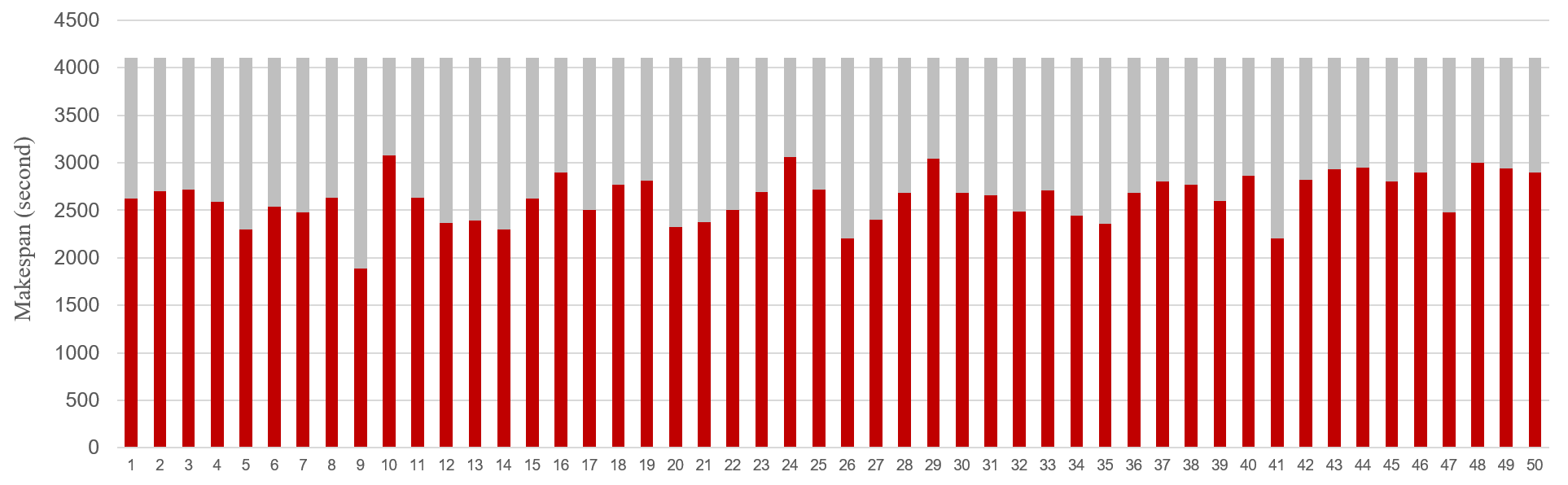}
	\caption{Test case two. Here, gray bars show the original run time and the red ones show our algorithms run time on a build.}
	\label{Test Case Two}
\end{figure}

In this test case, we state that our algorithm achieves improvements and produces reliable results. However, it does not generate exact the same results every time, but the run time values are very close to each other. In Figure \ref{Test Case Two}  gray bars has the same height, because it belongs to one build.

\subsubsection{Test Case Three}

\begin{figure}[h]
	\centering
	
	\includegraphics[scale=0.25]{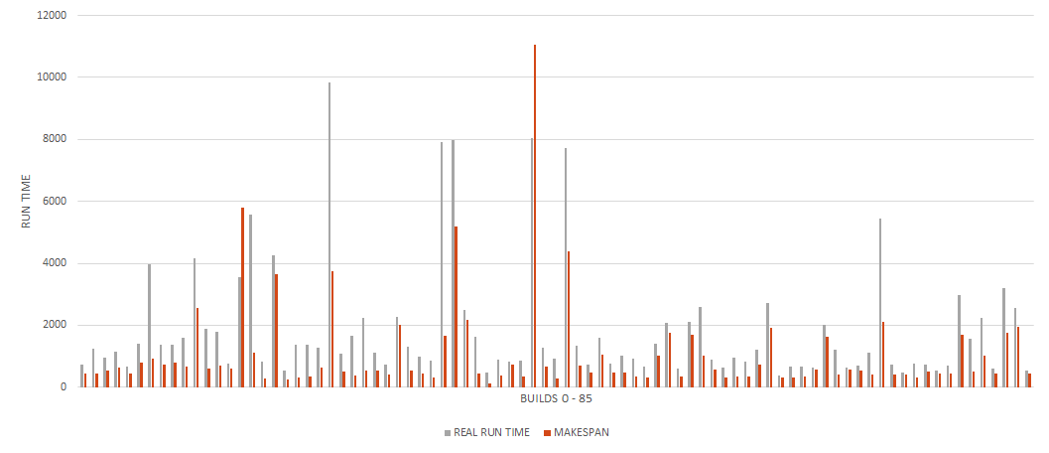}
	\caption{Test case three. Gray bars show the original run time as seconds and red bars shows makespan+run time of the proposed genetic algorithm.}
	\label{Test Case Three}
\end{figure}

Our aim is to find the best priority list when the job list is given to our algorithm as an input. The algorithm passes the priority list to the CI system as soon as it terminated to seek best priority list.
In the test case three, besides the total run time of a build, we also consider the total time consumed to find the best priority list. We aim to shorten this process. The run time of our algorithm exceeds the optimal duration, if it can not reach the target improvement. This would be the worst case for this problem. In Figure \ref{Test Case Three}  it is seen that there is still nearly \% 20 improvement on each build's run time. This means, summation of the makespan and our algorithm run time is still shorter than the user based created priority list.

\section{Conclusion}
Scheduling jobs in a smart way is a challenging problem. In this work we prepare two priority list generation algorithms by considering a smart fitness function. In this fitness function, we try to minimize makespan and machine count. As a result we reach 20\% improvement in our CI system. A unique chromosome representation method is presented in which there are machine information and job list. As a further work, we will introduce another crossover method which groups jobs according to their dependencies. With these improvements, the diversity of a population increases with a high probability and consumed time by crossover function will decrease.

\end{document}